\documentclass{article}

\usepackage[final]{neurips_2024}

\usepackage[utf8]{inputenc} %
\usepackage[T1]{fontenc}    %
\usepackage{hyperref}       %
\usepackage{url}            %
\usepackage{booktabs}       %
\usepackage{amsfonts}       %
\usepackage{nicefrac}       %
\usepackage{microtype}      %
\usepackage{xcolor}         %
\usepackage{authblk}
\usepackage{amsmath}
\usepackage{amsfonts}
\usepackage{amssymb}
\usepackage{textcomp}
\usepackage{algorithm}
\usepackage{algpseudocode}
\usepackage{graphicx}
\usepackage{siunitx}
\usepackage{shortcuts}
\usepackage{xcolor}
\usepackage{color}
\usepackage{authblk}
\definecolor{lightblue}{RGB}{173,216,230} %
\definecolor{lightgreen}{RGB}{144,238,144} %
\usepackage{wrapfig}

\title{TurboHopp: Accelerated Molecule Scaffold Hopping with Consistency Models}

\author[1]{\textbf{Kiwoong Yoo}}
\author[2]{\textbf{Owen Oertell}}
\author[3]{\textbf{Junhyun Lee}}
\author[1,3]{\textbf{Sanghoon Lee}}
\author[1,3]{\textbf{Jaewoo Kang}\thanks{Corresponding author.}\hspace{4pt}}

\affil[1]{AIGEN Sciences}
\affil[2]{Cornell University}
\affil[3]{Korea University}

\affil[ ]{}

\affil[ ]{\texttt{kiwoong.yoo@aigensciences.com, ojo2@cornell.edu, ljhyun33@korea.ac.kr, sanghoon.lee@aigensciences.com, kangj@korea.ac.kr}}

\begin{document}

\maketitle

\begin{abstract}

    Navigating the vast chemical space of druggable compounds is a formidable challenge in drug discovery, where generative models are increasingly employed to identify viable candidates. Conditional 3D structure-based drug design (3D-SBDD) models, which take into account complex three-dimensional interactions and molecular geometries, are particularly promising. Scaffold hopping is an efficient strategy that facilitates the identification of similar active compounds by strategically modifying the core structure of molecules, effectively narrowing the wide chemical space and enhancing the discovery of drug-like products. However, the practical application of 3D-SBDD generative models is hampered by their slow processing speeds. To address this bottleneck, we introduce TurboHopp, an accelerated pocket-conditioned 3D scaffold hopping model that merges the strategic effectiveness of traditional scaffold hopping with rapid generation capabilities of consistency models. This synergy not only enhances efficiency but also significantly boosts generation speeds, achieving up to 30 times faster inference speed as well as superior generation quality compared to existing diffusion-based models, establishing TurboHopp as a powerful tool in drug discovery. Supported by faster inference speed, we further optimize our model, using Reinforcement Learning for Consistency Models (RLCM), to output desirable molecules. We demonstrate the broad applicability of TurboHopp across multiple drug discovery scenarios, underscoring its potential in diverse molecular settings. The code is provided at \url{https://github.com/orgw/TurboHopp}
\end{abstract}

\section{Introduction}
The vast chemical space, with up to $10^{63}$ potential molecules (molecular weight under \SI{500}{\dalton}) (\cite{kirkpatrick2004chemical}), presents immense challenges in drug discovery, where the process often extends over a decade and incurs costs in the billions. Recent strides in molecular design have introduced advanced generative models that excel in de-novo generation and optimization within 3D molecular space (\cite{isert2023structure}). Despite these technological advances, challenges persist, such as ensuring synthesizability (\cite{gao2021amortized, fialkova2021libinvent}) and maintaining desirable drug-like properties. The scarcity of protein-ligand complex data further complicates these demanding issues. While 1D/2D models attempt to address these challenges though various methods (\cite{loeffler2024reinvent}), they often fall short in representing true 3D structures, and typically necessitating additional steps for proper conformation generation. On the other hand, 3D models (\cite{ragoza2022generating, luo20223d, baillif2023deep, peng2022pocket2mol}), which directly incorporate spatial structures, show greater promise in activity prediction regarding 3D structures but are hindered by significantly slower inference speeds (\cite{baillif2023deep}), which limit their practical applications such as fine-tuning via Reinforcement Learning (\cite{ding2024consistency, baillif2023deep, wallace2023diffusion}). Recent research efforts, such as those by \cite{lee2024fine}, have started fine-tuning 3D-autoregressive models with specialized reward functions to align with these stringent standards. Yet, notably absent are attempts involving Structure-Based Drug Design Diffusion Models(3D-SBDD DMs), as the prolonged inference times remain a formidable barrier, precluding the adoption of similar methodologies. In this work, we address this gap by successfully applying reinforcement learning to 3D-SBDD DMs, overcoming the challenge of prolonged inference times and demonstrating the feasibility of such approaches in practical drug design scenarios.

3D-SBDD-DMs (\cite{guan20233d, guan2024decompdiff, schneuing2022structure, torge2023diffhopp, igashov2024equivariant}) exhibit powerful generative capabilities but suffer from slow processing speeds due to their reliance on iterative sampling. Addressing these speed limitations is critical for several reasons. Primarily, faster processing speeds are essential for accelerating hit discovery phases, which could significantly increase the likelihood of identifying quality hits within a shorter time period. Additionally, the high computational demands of these models not only escalate costs but also limit access to advanced optimization techniques, thereby restricting the generation of molecules with desirable properties such as synthesizability, high binding affinity, and drug-likeness. Furthermore, real-time applications, such as interactive modeling sessions with human experts, require rapid model inference to provide effective feedback and facilitate iterative testing (\cite{sundin2022human}). By enhancing the speed of these models, it may be possible to enable real-time applications, allowing for on-the-fly modifications and adaptations during interactive sessions. Addressing the speed limitations of diffusion models is thus essential to fully realize these objectives and leverage the full potential of 3D-SBDD-DMs in practical applications.

To address slow inference of diffusion models, various strategies have been adopted to enhance their efficiency, in the computer vision domain such as improvement of solvers, optimization of diffusion processes, and model distillation techniques (\cite{cao2024survey}). The emergence of consistency models and its variants (\cite{song2023improved, ding2024consistency, song2023consistency}), heralds a rapid, efficient alternative capable of delivering high-quality results with significantly
reduced computational demands. However, these acceleration attempts have been minimally explored in the context of 3D drug design.

\begin{figure}[ht]
  \centering
    \includegraphics[width=0.8\textwidth]{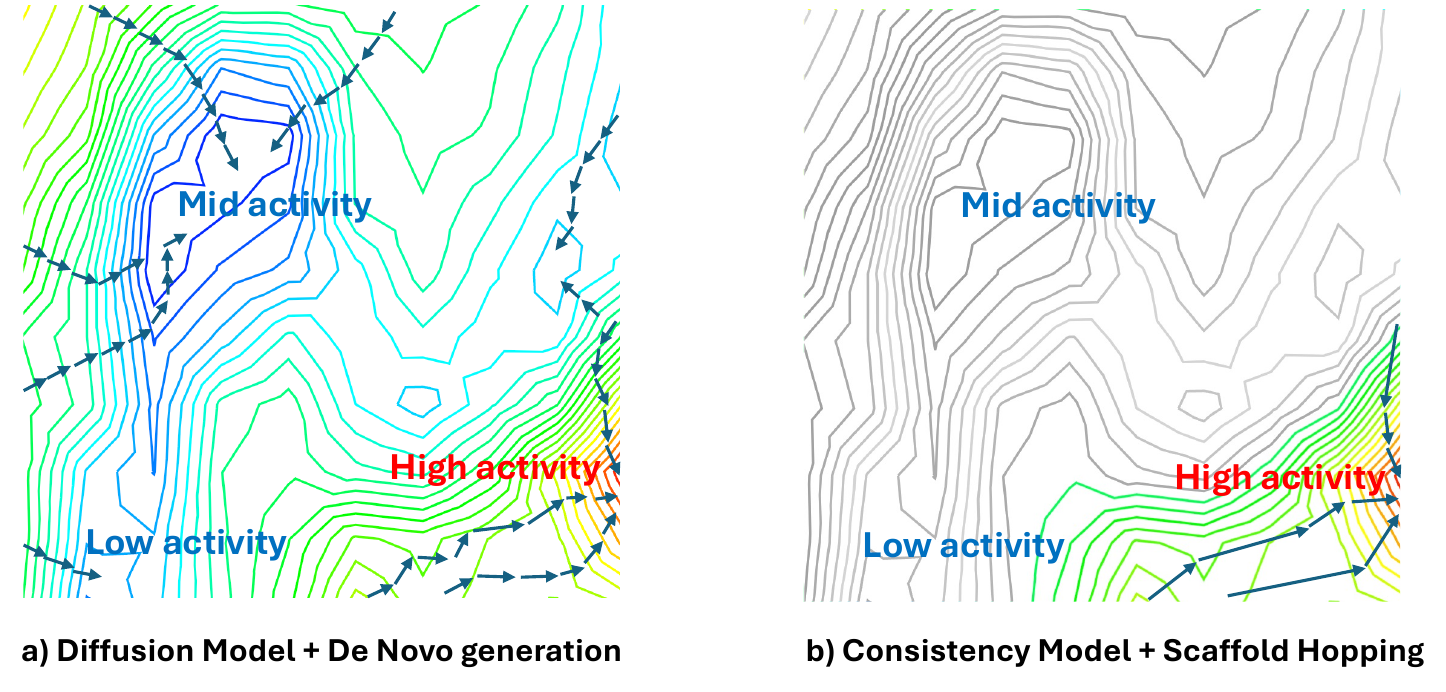}  %
    \caption{
    (a) Previous diffusion-based SBDD models methodically explore vast chemical space for pocket-active molecules, with short arrows symbolizing a gradual, step-wise inference process. (b) TurboHopp efficiently generates active ligands using a consistency model, which accelerates inference and reduces the number of steps, as illustrated by the longer arrows. Moreover, it strategically leverages the functional groups of high-activity reference molecules, shown as colored areas on the diagram, to optimize the exploration within targeted chemical space. }
    \label{fig:figure1}
    
\end{figure}

In this work, we present an efficient E(3)-equivariant consistency model that combines the rapid generation capabilities of consistency models with the effectiveness of scaffold hopping (\cref{fig:figure1}).  To substantiate the reliability and robustness of model performance, subsequent sections will detail results and comparisons with established diffusion and consistency models. Moreover, we incorporate reinforcement learning (RL) techniques (\cite{oertell2024rl}) to further refine our model based on objectives that are challenging to address through direct modeling alone, demonstrating its capability to surmount limitations encountered by previous 3D-SBDD models (\cite{harris2023posecheck}).

Our contributions are as follows:
\begin{itemize}
\item We propose a target-aware equivariant consistency model tailored for scaffold-hopping, achieving speeds up to 30 times faster than traditional DDPM-based models while achieving higher drug-likeness, synthesizability, connectivity and binding affinity.
\item By leveraging faster inference speeds, we apply Reinforcement Learning (RL) to enhance binding affinity scores and reduce steric clashes without redocking in 3D-SBDD—a strategy not previously explored with diffusion models. This approach enables targeted optimizations in molecule design and opens up opportunities to tackle metrics that were previously challenging to address through direct modeling.

\end{itemize}

\begin{figure}[ht]
  \centering
      \includegraphics[width=\textwidth]{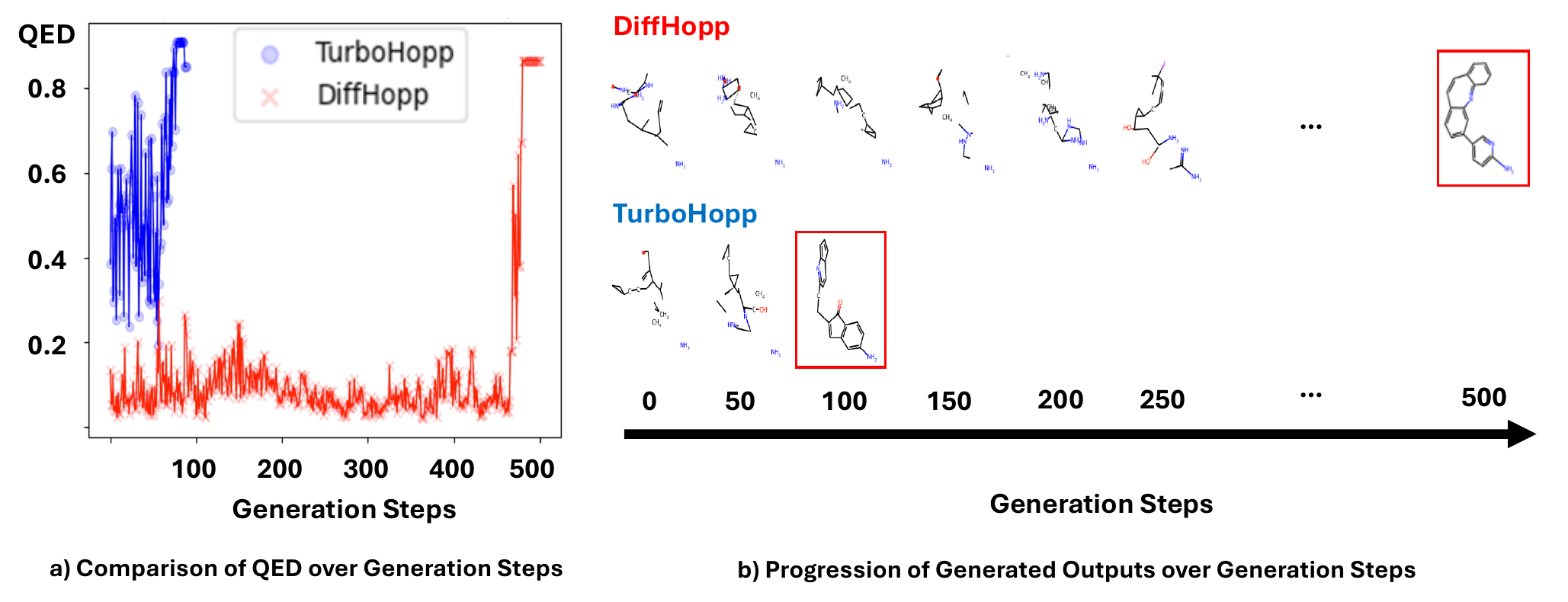}  
  \caption{(a) Comparison of QED scores, with TurboHopp (blue) reaching peak values faster compared to DiffHopp (red), throughout the generation process. (b) Progression of generated outputs of both models. Final steps are highlighted with red boxes.}
\end{figure}

\section{Related Works}
\subsection{Scaffold Hopping}
Scaffold hopping is a key strategy in drug discovery aimed at identifying new compounds that share biological activities but differ structurally from known molecules. This approach is crucial for enhancing drug properties and navigating intellectual property challenges. It includes techniques like heterocycle replacements, ring modifications, peptidomimetics for stability, and topology-based hopping, which innovates on molecular shape while maintaining activity (\cite{sun2012classification}). Deep-learning models, particularly in topology-based hopping, explore complex molecular structures to predict novel scaffolds, driving forward drug discovery by enabling the design of therapeutics with optimized efficacy and patentability (\cite{torge2023diffhopp}).

\subsection{Diffusion-Based Molecular Generative Models and Limitations}

Diffusion-based molecular generative models (DMs) apply noise processes and their reversal to generate novel molecular structures. These range from de-novo, non-conditional molecule creation to precision drug design within protein pockets, and extend to specific tasks like scaffold hopping and linker generation (\cite{hoogeboom2022equivariant, vignac2023midi, huang2023mdm, guan20233d, guan2024decompdiff, huang2024binding, torge2023diffhopp, igashov2022equivariant}). 

Despite their broad applicability, these models encounter substantial constraints.
First, the iterative reversal of noise through SDEs or ODEs often requires extensive computation, typically hundreds to thousands of iterations, resulting in slow model inference. This significantly constrains the speed at which new, active molecules can be identified, diminishing the practical efficacy of these models in time-sensitive applications (\cite{baillif2023deep}).
Second, diffusion models often produce molecular poses that fail to comply with key biophysical constraints, as highlighted by recent evaluations (\cite{harris2023posecheck}). \cite{guan20233d, guan2024decompdiff} struggle significantly in generating biophysically realistic molecular poses. Their strain energies are notably high, with median values around 1241.7 kcal/mol and 1243.1 kcal/mol, substantially exceeding the preferable baseline of 102.5 kcal/mol from the CrossDocked dataset (\cite{francoeur2020three}). Additionally, both models exhibit considerable issues with protein clashes: \cite{schneuing2022structure} shows a high rate of steric clashes with a mean score of 15.33 before redocking, and \cite{guan20233d} also has a problematic level, with a mean score of 9.08 prior to redocking. These high levels of steric clashes suggest that the molecules often have conformations that are too cramped or improperly aligned for effective protein binding, highlighting significant challenges in their ability to produce physically and biologically plausible interactions.

\subsection{Consistency Models}

Initially utilized in the vision domain, consistency models are either developed by distilling knowledge from existing diffusion models or trained as standalone entities. Defined as a distinct class of generative models (\cite{song2023consistency}), they address the computational inefficiencies of traditional diffusion models that require multiple iterative steps to remove noise and generate data. By directly transforming noise into data, consistency models significantly reduce generation time and offer the flexibility of multi-step sampling to improve output quality.

\subsection{RL for Consistency Models}

Reinforcement Learning for Consistency Models (RLCM) is a framework that enhances the speed and efficacy of training generative models through reinforcement learning (\cite{oertell2024rl}). It builds upon the fast inference speed of consistency models, enabling rapid training and inference cycles. RLCM is particularly useful as it fine-tunes models to optimize for specific reward functions, especially those hard to capture through direct modeling. In the molecule generation domain, to our knowledge there have been no attempts at fine-tuning diffusion or consistency models for task-specific purposes yet. By leveraging RLCM, we show promising evidence for mitigating issues in 3D-SBDD DMs (\cite{harris2023posecheck}), and believe that future iteration on the reward functions with our methods will enhance their practical applicability and effectiveness.

\section{Methods}

\subsection{Notations and Problem Definition}

\paragraph{Notations}  Atoms within proteins and ligands are represented by the graph \(G = \{h, x\}\), where \(h \in \mathbb{R}^{N \times F}\) are node features and \(x \in \mathbb{R}^{N \times 3}\) are the Cartesian coordinates of the atoms. The atomic elements are encoded as one-hot vectors within \(h\). We consider only those protein pocket atoms that are within an 8 Ångström radius of the ligand, simplifying the protein structure to a \(C_{\alpha}\) representation with node features \(h_P\) encoding residue types as one-hot vectors. Edges between atoms are defined for those within a 5 Ångström radius, incorporating both the distance \(d_{ij}\) and a normalized direction vector \(\frac{x_i - x_j}{d_{ij}}\).

\paragraph{Problem Definition} Given a context $u$ which integrates the functional groups of a molecule denoted as $g$, and a protein pocket $p$, our focus is to generate a scaffold, a key substructure of a molecule $Z$. A generative model, defined as $P_\theta(Z |u )$ maps uniform Gaussian noise to conditional probability distribution $Z$ given $u$. Ultimately, we aim to make sampling of scaffolds for any given molecular context $u$ within the set as efficient as possible, without losing overall quality. 
 
\paragraph{Dataset}
 We follow the dataset preprocessing scheme regarding filtering of compounds and determining of scaffolds as done in \cite{torge2023diffhopp}, filtering those above QED of 0.3, training on 19,378 protein-ligand complexes in PDBBind. We also adopt the same scaffold extraction method, using Murko-Bemis method (\cite{bemis1996properties}). 

\subsection{Background}

\paragraph{Diffusion Models}
In DDPMs, noise is incrementally added to the original data, represented as $\mathbf{Z}_0$, until it is completely transformed into a noise-dominated state $\mathbf{Z}_T$ after $T$ steps. This process is governed by an SDE:
\begin{equation}
    d\mathbf{Z} = \mathbf{f}(\mathbf{Z}, t) dt + \mathbf{u}(t) d\mathbf{w}
\end{equation}
where $\mathbf{w}$ represents standard Brownian motion, $\mathbf{f}(\mathbf{Z}, t)$ is a drift function, and $\mathbf{u}(t)$ is a diffusion coefficient associated with the noise process.
The reverse diffusion process is described by another SDE which incorporates the backward flow of time through $\mathbf{\tilde{w}}$, a Brownian motion in reverse, and modifies the forward SDE with a term involving the gradient of the log probability density.:
\begin{equation}
d\mathbf{Z} = [\mathbf{f}(\mathbf{Z}, t) - \mathbf{u}(t)^2 \nabla_{\mathbf{Z}} \log p_\theta(\mathbf{Z})] dt + \mathbf{u}(t) d\mathbf{\tilde{w}}
\end{equation}

The denoising process can be mathematically framed as an ordinary differential equation (ODE), also known as the Probability Flow ODE (\cite{song2020denoising}), whose solution trajectories have the same marginal density at time t:
\begin{equation}
    d\mathbf{Z} = [\mathbf{f}(\mathbf{Z},t) - \frac12 \mathbf{u}(t)^2 \nabla_{\mathbf{Z}} \log p_\theta(\mathbf{Z})] dt
\end{equation}
During training phase,  Diffusion Models (DMs) learn to estimate the score, given by \(\nabla_{\mathbf{Z}} \log p_\theta(\mathbf{Z})\) with a score model \(\mathbf{s}_\theta(\mathbf{Z}, t)\). Following the modifications by \cite{karras2022elucidating}, the drift term $\mathbf{f}(\mathbf{Z}, t)$ is set to zero and the diffusion term is defined as $\mathbf{u}(t) = \sqrt{2t}$. Consequently, the empirical Probability Flow ODE becomes:
\begin{equation}
    \frac{d\mathbf{Z}}{dt} = -t \mathbf{s}_\theta(\mathbf{Z}, t)
\end{equation}
This form is used to solve the ODE backwards in time, starting from a noise state $\mathbf{\tilde{Z}}$ sampled from $\mathcal{N}(0, t^2\mathbf{I})$. Using numerical ODE solvers like Euler and Heun, the model computes the trajectory $\{\mathbf{Z}_t\}_{t \in [0, T]}$, resulting in $\mathbf{\tilde{Z}}_0$ as an approximate reconstruction of the original data distribution.

\paragraph{Consistency Function}
We develop an equivariant consistency function \( f : (Z_t, t \mid u) \rightarrow Z_{\epsilon} \) for a solution trajectory \( Z_t \) of an ODE across \( t \in [\epsilon, T] \), where \(\epsilon\) is near zero. This function meets both self-consistency, ensuring \( f(Z_t, t \mid u) \) is uniform for all \( (Z_t, t) \) on the trajectory, and boundary conditions, mandating \( f(Z_{\epsilon}, \epsilon) = Z_{\epsilon} \) at the smallest time step \(\epsilon\). Successfully achieving these conditions guarantees that each solution \( Z_t \) can be traced back to its original state \( Z_0 \). The function maintains SE(3)-equivariance, adjusting centers as per methodologies outlined in \cite{xu2022geodiff}, and outputs both coordinates and node features.

\subsection{Model Architecture}
\begin{figure}[ht]
  \centering
    \includegraphics[width=1.0\textwidth]{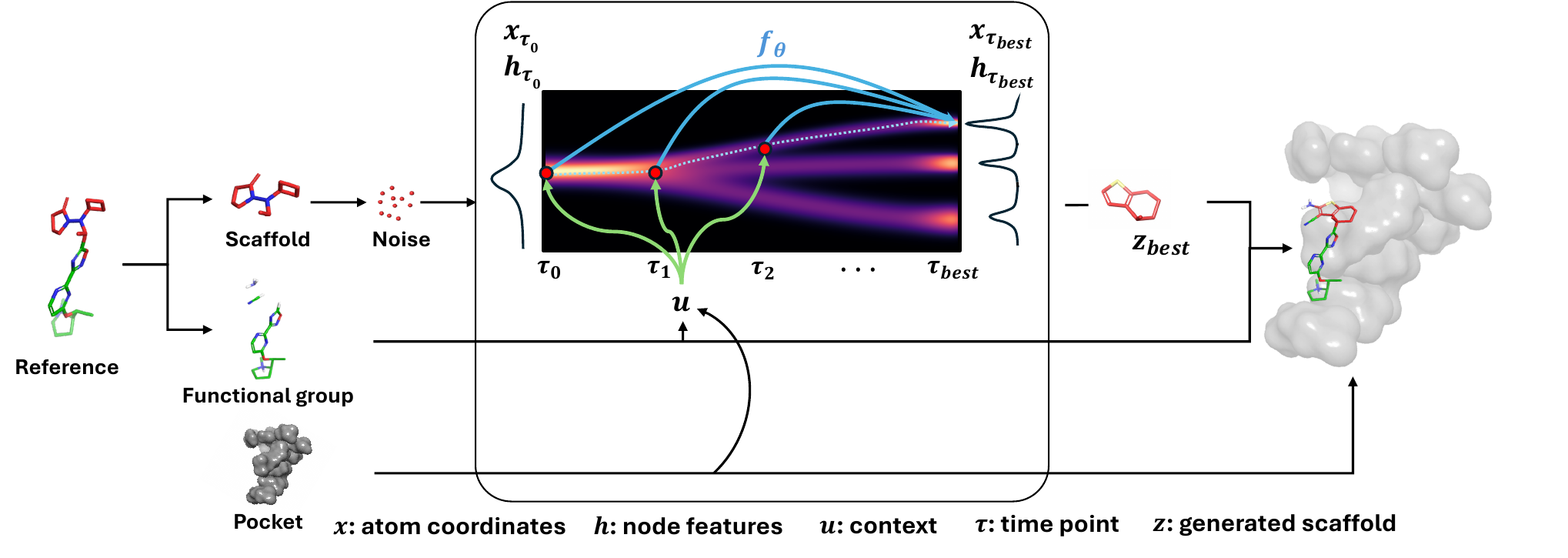}
  
    \caption{Model Architecture of \textcolor{lightblue}{TurboHopp}. Given a reference ligand and its corresponding protein pocket, an equivariant consistency model samples scaffolds conditioned on pocket substructure and functional groups. Models are trained to map points on the same PF-ODE path to the original data given \textcolor{lightgreen}{context}.}
\end{figure}
We wish to construct a new sample scaffold \( Z \) given a molecular context \( u \) ( \( u \) is the concatenation of the pocket \( p \) and functional groups \( g \)). 
An equivariant consistency function \( f_{\theta}(Z | u) \) is parameterized by a  free-form deep neural network $F_{\theta}$ which is a $(Z, t)$ related SE (3)-equivariant model. Theoretically, it can be any architecture that is SE(3)-equivariant. We use an adaptation of the equivariant Geometric Vector Perceptron (GVP) architecture (\cite{jing2020learning}). As done in (\cite{torge2023diffhopp}), all features are embedded into a shared feature space using separate MLPs. The time factor ${t}$ is embedded with a linear layer and added together as the input. As done in \cite{torge2023diffhopp},  hidden node features \( h' \) and \( x' \) are obtained though message-passing layers. The outputs of the consistency function are \( x'_t, h'_t = x'_z, \phi_{\text{out}} (h'_Z) \) where \( \phi_{\text{out}} \)is an MLP to map from embedding space to original data space.
As shown in \cref{eqn:5}, we separate coordinates and node features and parameterize the consistency model with a learnable function $f_{\theta}$ using skip connections.
\begin{align}
\label{eqn:5}
f_{\theta}(x_t, t \mid u) &= z_{\text{skip}}(t)x_t + z_{\text{out}}(t)x'_t & f_{\theta}(h_t, t \mid u) &= z_{\text{skip}}(t)h_t + z_{\text{out}}(t)h'_t
\end{align}
\begin{align}
\label{eqn:6}
z_{\text{skip}}(t) &= \frac{\sigma^2_{\text{data}}}{(t - \epsilon)^2 + \sigma^2_{\text{data}}} & z_{\text{out}}(t) &= \frac{\sigma_{\text{data}} (t - \epsilon)}{\sqrt{(t - \epsilon)^2 + \sigma^2_{\text{data}}}}
\end{align}
\( z_{\text{skip}}(t) \) and \( z_{\text{out}}(t) \) are differentiable time-dependent functions satisfying \( z_{\text{skip}}(\epsilon) = 1 \) and \( z_{\text{out}}(\epsilon) = 0 \), thus ensuring the function \( f_{\theta} \) adheres to the boundary condition and retains differentiability. Given \( \sigma_{\text{data}} \) as the standard deviation of data, when \( t = \epsilon \), setting \( z_{\text{skip}} (\epsilon) = 1 \) and \( z_{\text{out}} (\epsilon) = 0 \) ensures that the condition \( f_{\theta}(Z_{\epsilon}) = Z_{\epsilon} \) is naturally satisfied. The functions \( z_{\text{skip}} \), \( z_{\text{out}} \), and \( F_{\theta} \) are differentiable.
As stated in \cite{fan2023ec}, since ${z_t}$ and $F_{\theta}$ are both SE-(3) equivariant, and \( z_{\text{skip}}(t) \) and \( z_{\text{out}}(t) \) are invariant scalars , the SE (3)-equivariance is guaranteed.

\subsection{Model Training}

 We apply \cref{alg:training_alg} for training our consistency model. Scaffold \( Z \) extracted from the dataset is segmented into its coordinate and feature components, \( x \) and \( h \), respectively. Next, adjustments are made to the coordinates and features along the ODE trajectory \( \{x_t\}_{t \in [\epsilon, T]}\) and \( \{h_t\}_{t \in [\epsilon, T]}\), respectively, by incorporating perturbations \( {\epsilon}_x \) and \( {\epsilon}_h \). The loss function \( L(\theta) \) reduces discrepancies between successive time steps \( f_{\theta}(Z_{n+1}, t_{n+1} | u) \) and \( f_{\theta}(Z_{n}, t_{n} | u) \), ensuring self-consistency. 

 Let $f_{\theta}^{n,x}$ and $f_{\theta}^{n,h}$ denote the coordinate and feature components of $f_{\theta}(Z_n, t_n | u)$, respectively. Similarly, let $f_{\theta}^{n+1,x}$ and $f_{\theta}^{n+1,h}$ denote the coordinate and feature components of $f_{\theta}(Z_{n+1}, t_{n+1} | u)$. The loss function $L(\theta)$ is defined as:

\begin{equation}
L(\theta) = \text{MSE}(f_{\theta}^{n+1,x}, f_{\theta}^{n,x}) + \text{MSE}(f_{\theta}^{n+1,h}, f_{\theta}^{n,h})
\end{equation}

To enhance the stability of the training process for \( f_{\theta} \), a secondary function \( f_{\theta^-} \), which utilizes parameters \( \theta^- \) as the exponential moving average (EMA) of \( \theta \), is employed. This methodology effectively reduces the variability between consecutive predictions by accounting for incremental perturbations, as detailed in \cite{song2023improved}. The combined loss function measures the mean squared error (MSE) between adjacent states for both coordinates \( x \) and features \( h \):

Parameter updates are conducted using stochastic gradient descent for \( \theta \), while \( \theta^- \) is incrementally adjusted using an EMA protocol where \( \mu \) indicates the decay rate, enhancing the model’s responsiveness and accuracy.
\begin{align}
L(\theta, \theta^-) &= \text{MSE}(f_{\theta}^{n+1,x}, f_{\theta^-}^{n,x}) + \text{MSE}(f_{\theta}^{n+1,h}, f_{\theta^-}^{n,h}) \\
\theta^- &= \mu \theta^- + (1 - \mu)\theta
\end{align}

\subsection{Reinforcement Learning Architecture }

We adopt RLCM from \cite{oertell2024rl}, where we biject a Markov decision process to multistep consistency model inference. More formally, we adopt this mapping:
\begin{align*}
    \bs_t &\defeq (\wt Z_{\tau_t}, \tau_t, \bu) &&\pi(\ba_t | \bs_t) \defeq f_{\theta}\left(\wt Z_{\tau_t}, \tau_t | \bu \right) + \eps &&P(\bs_{t+1} | \bs_t, \ba_t) \defeq (\delta_{\wt Z_{\tau_{t+1}}}, \delta_{\tau_{t+1}}, \delta_{\bu})\\
    \ba_t &\defeq \wt Z_{\tau_{t+1}} &&\mu \defeq \left(\Ncal(0, I), \delta_{\tau_0}, p(\bu)\right) &&R_H(\bs_H) = r(f_\theta(\wt Z_{\tau_H}, \tau_H, \bu), \bu)
\end{align*}

where $\eps$ represents the noise added during the noising procedure of the consistency model.

As with \cite{oertell2024rl}, we create a mapping from MDP states to noised states along the consistency model trajectory. Likewise, we follow a deterministic transition from the state to the predicted action. This action is a combination of the output from the consistency model convolved with random noise in accordance with the DDPM stochastic differential equation (\cite{ho2020denoising, song2021scorebased}). Because this policy now becomes stochastic, we can apply \cref{alg:pg-rlcm} (a modified version of reinforce with a clipping). The agent then receives a reward at the end of the trajectory corresponding to a combination of desirable chemical attributes. We use PPO (\cite{schulman2017proximal}) for the core optimization strategy but others like REBEL (\cite{gao2024rebel}) have been shown to work for RLCM optimization with sometimes stronger results.

\subsection{Metric-based Sampling}

By leveraging the inherent efficiency of consistency models, we tailor molecule designs to precise standards. Contrary to the fixed-step approach suggested by \cite{song2023consistency}, our strategy utilizes a flexible, customizable metric that adapts to the variability in protein conditions, as evidenced by our empirical findings shown in \cref{fig:turbohopp_mols}. This approach allows for dynamic adjustment during generation, ensuring optimal results by selecting scaffolds with the highest evaluated scores. This not only boosts efficiency but also significantly enhances molecule properties, achieving superior docking and QED scores that surpass those found in the initial dataset.

In \cref{alg:sampling_alg}, initial samples are drawn from random Gaussian noise \(\widetilde{Z}_T \sim \mathcal{N}(0, T^2I)\).\(f_{\theta}\) is then applied to generate initial scaffolds from these samples: \(\widetilde{Z}_t = f_{\theta}(Z_T, T)\). Across a sequence of time points \( \{\tau_1, \tau_2 \ldots \tau_{N-1}\}\) we iteratively refine the generated scaffold. After reaching a designated step m, we score metrics to assesses the quality of each \(Z_t\). Then we select the one with the best metric at the end of all the iterations as the final output. Our metric function is based on drug-likeness and synthesizability.

\section{Experiments and Results}

\begin{table}[t]
  \centering
  \caption{\textbf{DiffHopp VS TurboHopp VS TurboHopp Metric Sampling} .Results on models trained on PDBBind. Mean and standard deviation of the common molecular metrics for the molecules from  the baseline models as well as time-step variations of our model."metric" refers that inference was done with metric-based sampling. QVina score (kcal/mol) refers to estimated binding affinity measured by QVina2. Time refers to average time (seconds) required to generate a batch of molecules per complex. Best metrics are in bold. 2nd best are underlined.}
  \vspace{10pt} 
  \label{table:metric_turbohopp}
  \makebox[0pt]{%
  {\scriptsize %
  \begin{tabular}{lcccccccc}
    \toprule
    Method & Connectivity (↑) & Diversity (↑) & Novelty (↑) & QED (↑) & SA (↑) & QVina (↓) & Steps & Time \\
    \midrule
    DiffHopp & 0.918\tiny{$\pm$}0.23 & 0.589\tiny{$\pm$}0.17 & \underline{0.999\tiny{$\pm$}0.01} & 0.621\tiny{$\pm$}0.12 & 0.662\tiny{$\pm$}0.14 & -7.923\tiny{$\pm$}3.29 & 500 & 107.10 \\
    DiffHopp\textsubscript{EGNN} & 0.752\tiny{$\pm$}0.41 & \textbf{0.641\tiny{$\pm$}0.15} & \textbf{1.000\tiny{$\pm$}0.01} & 0.510\tiny{$\pm$}0.12 & 0.592\tiny{$\pm$}0.11 & -7.210\tiny{$\pm$}1.45 & 500 & - \\

    DiffHopp\textsubscript{metric} & 0.927\tiny{$\pm$}0.26 & 0.597\tiny{$\pm$}0.20 & \textbf{1.000\tiny{$\pm$}0.02} & 0.634\tiny{$\pm$}0.18 & 0.670\tiny{$\pm$}0.13 & -7.783\tiny{$\pm$}2.78 & 500 & 442.48 \\

    \midrule
    
    TurboHopp-50 & 0.872\tiny{$\pm$}0.19 & {0.562\tiny{$\pm$}0.18} & \textbf{1.000\tiny{$\pm$}0.00} & 0.576\tiny{$\pm$}0.12 & 0.635\tiny{$\pm$}0.25  & -7.823\tiny{$\pm$}1.53 & 50 & \textbf{3.19} \\
    
    TurboHopp-100 & {0.948\tiny{$\pm$}0.22} & 0.563\tiny{$\pm$}0.23  & {0.997\tiny{$\pm$}0.09} & {0.589\tiny{$\pm$}0.19} & {0.724\tiny{$\pm$}0.12} & -8.272\tiny{$\pm$}1.21 & 100 & 5.69 \\
    
    TurboHopp-150 & \textbf{1.000\tiny{$\pm$}0.00} & \underline{0.573\tiny{$\pm$}0.12} & \textbf{1.000\tiny{$\pm$}0.00} & {0.618\tiny{$\pm$}0.19} & 0.715\tiny{$\pm$}0.13 & \underline{-8.277\tiny{$\pm$}1.87} & 150 & 15.82\\
    \midrule
    
    TurboHopp-50\textsubscript{metric} & 0.917\tiny{$\pm$}0.28 & \underline{0.612\tiny{$\pm$}0.21} & \textbf{1.000\tiny{$\pm$}0.00} & 0.583\tiny{$\pm$}0.20 & 0.631\tiny{$\pm$}0.15  & -7.865\tiny{$\pm$}1.72 & 50 & \underline{3.83} \\
    
    TurboHopp-100\textsubscript{metric} & \underline{0.997\tiny{$\pm$}0.06} & 0.561\tiny{$\pm$}0.22  & \textbf{1.000\tiny{$\pm$}0.00} & \textbf{0.664\tiny{$\pm$}0.19} & \textbf{0.737\tiny{$\pm$}0.11} & \textbf{-8.319\tiny{$\pm$}1.38} & 100 & 7.14\\
    
    TurboHopp-150\textsubscript{metric} & \textbf{1.000\tiny{$\pm$}0.00} & 0.569\tiny{$\pm$}0.22 & \textbf{1.000\tiny{$\pm$}0.00} & \underline{0.648\tiny{$\pm$}0.19} & \underline{0.727\tiny{$\pm$}0.11} & {-8.219\tiny{$\pm$}1.722} & 150 & 20.88\\
    \midrule
    PDBBind Test  & 1.000\tiny{$\pm$}0.00 & - & 1.000\tiny{$\pm$}0.00 & 0.599\tiny{$\pm$}0.17 & 0.742\tiny{$\pm$}0.11 & -8.712\tiny{$\pm$}0.18 \\
    \bottomrule
  \end{tabular}
  }}
\end{table}

\begin{table}[t]
  \centering
  \caption{\textbf{TurboHopp VS TurboHopp-RLCM with Metric Sampling}. We optimized TurboHopp-100 with RLCM using $r_{\text{docking score}}$. Single optimization task is based on multiple reference protein-ligand complexes of the PDBBind testset; we do not optimize one protein at a time. To fully support faster parallel multi ligand - multi protein docking, we used AutoDockGPU instead of QVina2. Binding affinity scores surpassed reference docking scores without losing drug-likeliness and synthesizability.}
  \vspace{10pt} 
  \label{table:metric_rlcm}
  \makebox[0pt]{%
  {\scriptsize %
  \begin{tabular}{lcccccccc}
    \toprule
    Method & Connectivity (↑) & Diversity (↑) & Novelty (↑) & QED (↑) & SA (↑) & Vina (↓) & Steps & Time \\
    \midrule
    
    TurboHopp-100\textsubscript{metric} & \textbf{0.997}\tiny{$\pm$}0.06 & 0.561\tiny{$\pm$}0.22  & \textbf{1.000}\tiny{$\pm$}0.00 & \textbf{0.664}\tiny{$\pm$}0.19 & \textbf{0.737}\tiny{$\pm$}0.11 & -8.298\tiny{$\pm$}1.82 & 100 & 7.14\\
    
    TurboHoppRL-50\textsubscript{metric} & 0.980\tiny{$\pm$}0.14 & \textbf{0.869\tiny{$\pm$}0.17} & 0.936\tiny{$\pm$}0.17 & 0.619\tiny{$\pm$}0.18 & 0.680\tiny{$\pm$}0.13  & \textbf{-9.804\tiny{$\pm$}2.84}  & 50 & \textbf{3.69} \\
    \midrule

    PDBBind Test & 1.000\tiny{$\pm$}0.00 & - & 1.000\tiny{$\pm$}0.00 & 0.599\tiny{$\pm$}0.17 & 0.742\tiny{$\pm$}0.11 & -8.643\tiny{$\pm$}2.08 \\
    \bottomrule
  \end{tabular}
  }}
\end{table}

\subsection{Performance Comparison with DDPM-based baseline}
\label{section:ddpm comparison}
In this section, we compare 3D scaffold generation qualities between TurboHopp and DiffHopp (\cite{torge2023diffhopp}), a DDPM-based scaffold-hopping model. For TurboHopp, we also vary the number of steps in multi-step generation.
\paragraph{Evaluation Metrics}

We evaluate sampling quality using several metrics outlined in \cite{torge2023diffhopp} . Drug-likeness and synthetic accessibility are assessed with  \cite{bickerton2012quantifying} and \cite{ertl2009estimation} respectively. Connectivity is defined as the percentage of fully connected molecules without fragments. Diversity is calculated using Tanimoto Dissimilarity, according to \cite{alhossary2015fast}. Novelty is determined by the proportion of molecules that were not present in the training dataset. Additionally, we use Vina Score from QVina2 \cite{alhossary2015fast} to estimate binding affinity. 

\paragraph{Results and Discussion}

Under the same context conditions, for a batch of 150 molecules, TurboHopp generates on average 5-30 times faster than baseline model (\cite{torge2023diffhopp}). Furthermore, as shown in \cref{table:metric_turbohopp}, compared to DMs, performance is superior in all metrics except diversity, which most likely falls compared to DMs due to the fact that consistency models are trained to have consistent results for data on the same trajectory. Also as total steps for multi-step generation increases, we can see an increase in diversity, due to outputs being refined less as steps decrease.

We applied the same score-based sampling method to the baseline model (DiffHopp\textsubscript{Scored}) and compared to the model without using it (DiffHopp). All metrics rose except binding affinity. However, in DiffHopp, this method significantly reduces sampling efficiency, as it necessitates more frequent evaluations of the generated products towards the end. In contrast, TurboHopp had a major increase in all metrics with a small downside of increased generation time of only a few seconds. This is possible due to the fact that there are less steps required to generate during multi-step generation, thus less steps to evaluate afterwards. This illustrates how, with the right methods, efficient sampling can significantly enhance the search for high-quality molecules within chemical space.

\subsection{Performance Optimization using RLCM}

In this section, we describe the optimization of our model using two specifically designed reward functions, each aimed at achieving distinct objectives in molecular design:
\begin{itemize}
    \item \textbf{Binding Affinity:} To enhance binding affinity, we began training with TurboHopp-100 and reduced the sampling steps to 50. The reward function is defined as:
    \begin{equation}
    r_{\text{docking score}}(s) = -2 \left\{ DS(s) - RS(s) \right\} + QED(s) + \frac{10 - SA(s)}{9} + C(s)
    \end{equation}
    where \(DS(s)\) and \(RS(s)\) represent the docking scores of the generated and reference molecules, respectively, highlighting the importance of relative score differences during learning. \(SA(s)\) signifies synthetic accessibility, and \(C(s)\) denotes connectivity. Following the insights from \cite{lee2023exploring}, we incorporate \(QED\) and \(SA\) to mitigate reward hacking, with connectivity included to ensure the structural validity of the molecules.
    
    \item \textbf{Protein Steric Clashes:} Steric clashes are defined as interactions where the distance between a protein and ligand atom is less than their combined van der Waals radii by a tolerance of $0.5$ \AA, indicating an energetically unfavorable and often physically implausible pose, crucial for SBDD performance evaluation. We refer to \cite{harris2023posecheck} for estimating number of clashes. To robustly manage steric clashes, we retained the original 100 steps from TurboHopp-100. The corresponding reward function is:
    \begin{equation}
    r_{\text{steric clash}}(s) = C(s) + QED(s) - \left(SC(s) - RSC(s)\right)
    \end{equation}
    where \(SC\) and \(RSC\) denote the steric clash scores for the generated and reference molecules, respectively, with the difference between these scores included to weigh their importance accurately.
\end{itemize}

For both models, we initially optimize QED and Connectivity for 50 and 100 steps, respectively, and subsequently finetune each task with the defined rewards metrics. \cref{table:metric_rlcm} shows that though metric-based sampling and RLCM, docking scores surpass those of the training dataset.

Unlike in the reward conditioned image generation setting (\cite{fan2023DPOK,black2024ddpo}), we omit any KL regularization to the base model during training. This decision stems from the nature of drug discovery where overfitting is less of a concern as long as the model maintains high scores in targeted areas. Furthermore, we empirically found that our approach does not significantly compromise diversity and novelty.

\paragraph{Optimization Metrics}

We optimize TurboHoppRL for the same test set we used for evaluating TurboHopp (refer to \cref{table:metric_rlcm}). Optimization encompasses QED, synthesizability, connectivity, docking score and number of protein clashes.  For training stability and computational efficiency, we utilize AutodockGPU (\cite{tang2022accelerating}) to estimate binding affinities, replacing  QVina2 (\cite{alhossary2015fast}) used for \cref{section:ddpm comparison} .

\begin{wrapfigure}{r}{0.6\textwidth}
\centering
\includegraphics[width=0.6\textwidth]{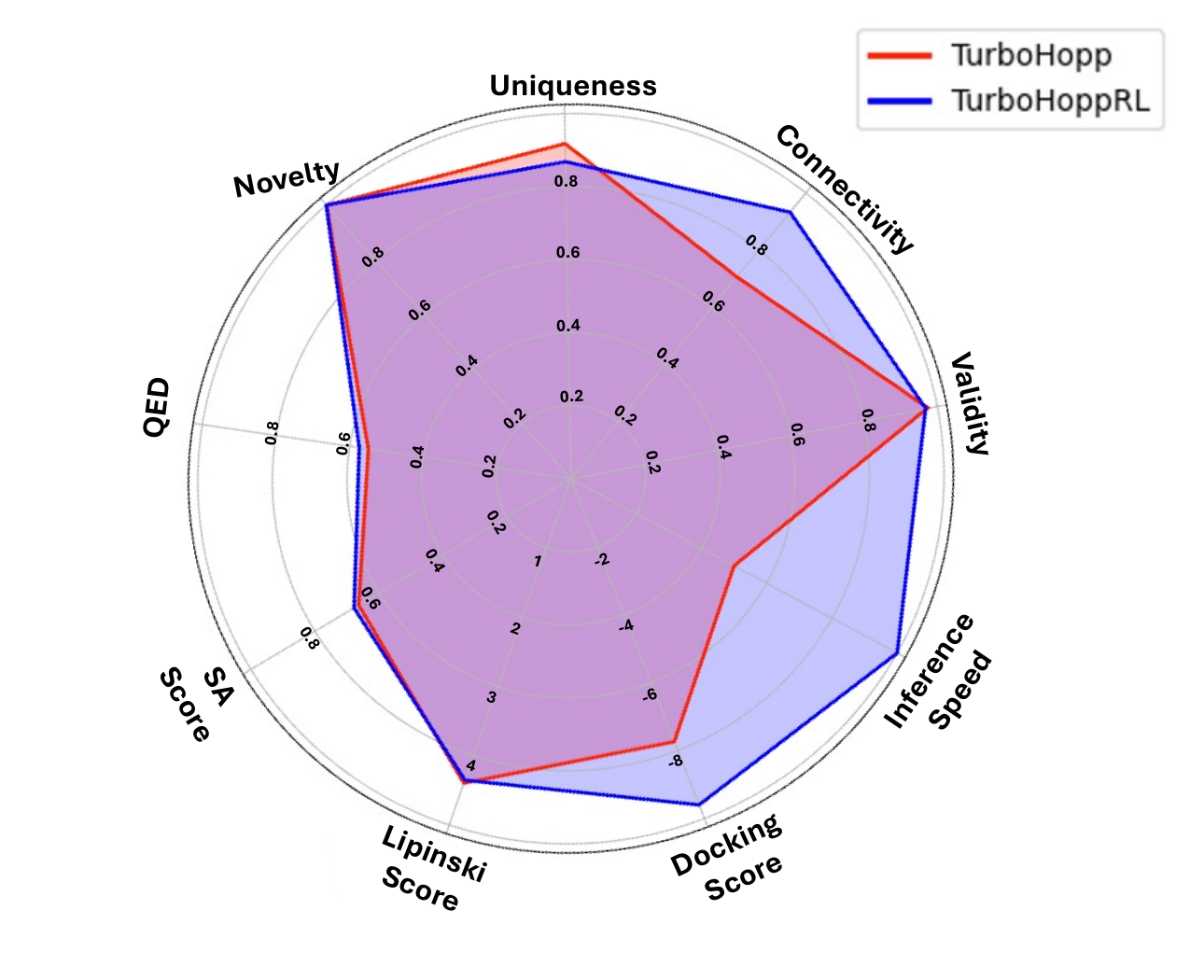}
\caption{Compared to TurboHopp-100 (Red), TurboHoppRL-50 (Blue) has enhancing binding affinity without losing performance in other metrics.}
\label{fig:turbohopprl_docking}
\vspace{-20px}
\end{wrapfigure}

\cref{fig:turbohopprl_docking} and \cref{table:metric_rlcm} compares TurboHopp-100 and TurboHoppRL-50. During training, we achieved higher binding affinity scores of about 2-3 kcal/mol compared to TurboHopp-100. Examples can be found at \cref{fig:docking-examples}.

As highlighted in \cref{fig:turbohopp-clashes}, we also fine-tuned a model to minimize steric clashes, achieving promising results as shown in \cref{fig:turbohopp-clashes_example}. However, optimizing the reward function further could further enhance clash minimization and improve the balance between metrics, which we propose as a direction for future research. 
While RL fine-tuned models are not necessarily expected to generalize to unseen molecules in drug discovery, there is potential for some generalization capabilities (\cite{oertell2024rl}), a topic also reserved for future exploration.

\begin{table}[t]
  \centering
  \caption{\textbf{TurboHopp VS 3D-SBDD inpainting models}. Results on models trained on CrossDocked. "Inpainting" refers to models using inpainting method explained in Algorithm 1. "metric" refers that inference was done with metric-based sampling. QVina score (kcal/mol) refers to estimated binding affinity measured by QVina2. Time refers to average time (seconds) required to generate a batch of molecules per complex. Best metrics are in bold. 2nd best are underlined. }
  \vspace{10pt} 
  
  \label{table:metrics_sbdd}
  \makebox[0pt]{%
  {\scriptsize %
  \begin{tabular}{lccccccccc}
    \toprule
    Method & Validity(↑) & Connectivity (↑) & Diversity (↑) & Novelty (↑) & QED (↑) & SA (↑) & QVina (↓) & Time \\
    \midrule
    
    TargetDiff\textsubscript{inpainting} & 0.927\tiny{$\pm$}0.13 & 0.826\tiny{$\pm$}0.12 & \underline{0.841\tiny{$\pm$}0.13} & 0.914\tiny{$\pm$}0.08 & 0.424\tiny{$\pm$}0.11 &  0.661\tiny{$\pm$}0.14 & -5.896\tiny{$\pm$}1.93 & 740.33 \\
    
    DecompDiff\textsubscript{inpainting} & 0.876\tiny{$\pm$}0.33 & 0.722\tiny{$\pm$}0.17 & \textbf{0.856\tiny{$\pm$}0.21} & 0.895\tiny{$\pm$}0.18 & 0.420\tiny{$\pm$}0.07 &  0.648\tiny{$\pm$}0.15 & -6.225\tiny{$\pm$}0.87 & 1263.72 \\

    \midrule
    
    TurboHopp-100 & 
    \underline{0.990\tiny{$\pm$}0.10} & \underline{0.853\tiny{$\pm$}0.36}  & {0.484\tiny{$\pm$}0.27} & 
    \textbf{0.936\tiny{$\pm$}0.24} & 
    \underline{0.488\tiny{$\pm$}0.21} & 
    \underline{0.702\tiny{$\pm$}0.13} & 
    \underline{-7.051\tiny{$\pm$}2.14} & 
    \textbf{6.17} \\
    
    TurboHopp-100\textsubscript{metric} & \textbf{0.993\tiny{$\pm$}0.08} & 
    \textbf{0.906\tiny{$\pm$}0.29}  & 
    0.486\tiny{$\pm$}0.28  & 
    \textbf{0.935\tiny{$\pm$}0.25} & \textbf{0.502\tiny{$\pm$}0.21} &
    \textbf{0.710\tiny{$\pm$}0.14} &
    \textbf{-7.204\tiny{$\pm$}2.00} & 
    \underline{8.18}\\
    \midrule  

    CrossDocked Test & 1.000\tiny{$\pm$}0.00 & - & 1.000\tiny{$\pm$}0.00 & 0.599\tiny{$\pm$}0.17 & 0.476\tiny{$\pm$}0.21 & 0.727\tiny{$\pm$}0.14 &
    -7.510\tiny{$\pm$}2.36\\
    \bottomrule
  \end{tabular}
  }}
\end{table}

\subsection{Performance Comparison with diffusion-based 3D-SBDD inpainting models}

Our model is compared with other diffusion based 3D-SBDD models, \cite{guan20233d} and \cite{guan2024decompdiff}. Inpainting was applied (\cite{lugmayr2022repaint}, {\cref{alg:inpainting}) on recent conditional de-novo diffusion models to create variations suitable for fair comparison. The same Bemis-Murcko scaffold determines scaffolds and functional groups. With regards to sampling conditions, the number of atoms is fixed to the reference scaffold for all models. For \cite{guan2024decompdiff}, since it additionally uses bond diffusion, a bond mask was created accordingly, and reference priors are used while sampling. The sampling hyperparameters for inpainting (resampling and jump length parameters) were determined by finding the ones with best validity.
Also, since these models were trained on CrossDocked, our model was additionally trained on CrossDocked for fair comparison. The same train-test split suggested in \cite{guan2024decompdiff} is followed, with an additional QED minimum filter of 0.3 when constructing the dataset. Only the alpha carbon residues of the protein pocket atoms are used to reduce computational burden.

\paragraph{Results and Discussion}

\cref{table:metrics_sbdd} shows that despite having lower diversity, our model has much faster generation speed as well as relatively high docking score compared to inpainting versions of reference SBDD models. Our findings show that a custom scaffold-hopping model outperforms a repurposed de-novo model.

Geometric properties and ring distributions ({\cref{alg:geometric_props}}) were also compared. Results show that our model has closer bond length/angle/atom-atom length distributions to the reference molecules compared to \cite{guan20233d}, but poorer results compared in bond/torsion angles. In all aspects, \cite{guan2024decompdiff} was outstanding largely because it learns the distribution of bonds while ours do not. For ring distributions, results show that our model is capable of generating similar ring types compared to the reference. In future research, we plan to design our model to learn bond properties as well.

\section{Conclusion}

We introduced TurboHopp, an equivariant conditional consistency model that significantly boosts generation efficiency and outperforms traditional DDPM-based models in various metrics. Empirical results indicate that our model achieves a 5-30 times improvement in generation speed while maintaining or enhancing traditional quality metrics. Additionally, by leveraging this increased inference speed, we demonstrated that reinforcement learning (RL) is feasible for 3D-SBDD-DMs, enabling the fine-tuning of generative models to meet specific requirements. Most importantly, our method has the potential to be adapted to numerous variations explored within 3D-SBDD-DMs, increase speed and efficiency.

Key areas for improvement and future research include enhancing the denoising model, expanding dataset size, and diversifying reward functions for drug discovery tasks. Incorporating bond diffusion (\cite{peng2023moldiff}) and including hydrogens in proteins and molecules could significantly bolster the robustness of the denoising model. In the consistency model, techniques such as eliminating the Exponential Moving Average (EMA), adopting Pseudo-Huber loss, and refining noise and loss schedules may yield more robust outcomes (\cite{song2023improved}). Furthermore, for RLCM, exploring reward functions that integrate steric energies and interaction fingerprints presents a promising avenue for future research.

\section*{Acknowledgements}

This work was supported in part by the National Research Foundation of Korea [NRF-2023R1A2C3004176]; the Ministry of Health \& Welfare, Republic of Korea [HR20C0021(3)]; the ICT Creative Consilience program through the Institute of Information \& Communications Technology Planning \& Evaluation (IITP) grant funded by the MSIT [IITP-2024–2020-0-01819];  the Ministry of Health \& Welfare and Ministry of Science and ICT, Republic of Korea [RS-2024-00462471]; the Artificial intelligence industrial convergence cluster development project funded by the Ministry of Science and ICT (MSIT, Korea) \& Gwangju Metropolitan City; and the Information and Communications Promotion Fund  through the National IT Industry Promotion Agency (NIPA), funded by the Ministry of Science and ICT (MSIT), Republic of Korea.

\bibliographystyle{unsrtnat}
\bibliography{mybib} %

\appendix
\newpage
\section{Algorithm Pseudocode}
In this section we present the pseudocode for consistency model training (\cref{alg:training_alg}), consistency model sampling (\cref{alg:sampling_alg}), RL training (\cref{alg:pg-rlcm}), and inpainting. (\cref{alg:inpainting})
\begin{algorithm}[ht]
\caption{Training}
\label{alg:training_alg}
\begin{algorithmic}[1]
\State \textbf{Input:} dataset $D = \{(Z_i, u_i)\}_{i \in M}$, where $M$ refers to the number of data points in $D$, $Z_i$ is composed of $x_i$ and $h_i$, initial model parameter $\theta$, learning rate $\eta$, step schedule $\mathcal{N}(\cdot)$, EMA decay rate schedule $\mu(\cdot)$, $\theta^- \leftarrow \theta$ and $k \leftarrow 0$;

\Repeat
    \State Sample $z, u \sim D$, and $n \sim \mathcal{U}[1, N(k) - 1]$;
    \State Decompose $z$ into $x$ and $h$;
    \State Sample ${\epsilon}_v \sim \mathcal{N}(0, I)$ for $v \in \{x, h\}$;
    \State Subtract center of gravity from ${\epsilon}_x$;
    \State Define $f_{\theta}^{n,v} \leftarrow f_{\theta}(v + t_n \cdot {\epsilon}_v, t_n | u)$ for $v \in \{x, h\}$;
    \State $L(\theta, \theta^-) \leftarrow \sum_{v \in \{x, h\}} \text{MSE}(f_{\theta}^{n+1,v}, f_{\theta}^{n,v})$;
    \State $\theta \leftarrow \theta - \eta \nabla_{\theta} L(\theta, \theta^-)$;
    \State $\theta^- \leftarrow \mu(k) \theta^- + (1 - \mu(k))\theta$;
    \State $k \leftarrow k + 1$;
\Until{convergence}
\end{algorithmic}
\end{algorithm}

\begin{algorithm}[ht]
\caption{Sampling with Scoring and Selection}
\label{alg:sampling_alg}
\begin{algorithmic}[1]
\State \textbf{Input:} Consistency model \(f_{\theta}(\cdot, \cdot)\) = (\(f_{\theta}^{x}\),\(f_{\theta}^{h}\)), sequence of time points \( \{ \tau_1, \tau_2, \ldots, \tau_{N-1} \} \) where \( \tau_1 > \tau_2 > \ldots > \tau_{N-1} \), evaluation starting point \(m\), where \(1 \leq m \leq N-1\)

\State Sample ${\epsilon}_x \sim \mathcal{N}(0, I)$ and ${\epsilon}_h \sim \mathcal{N}(0, I)$
\State Subtract center of gravity from ${\epsilon}_x$
\State ${\epsilon} \leftarrow [\epsilon_x, \epsilon_h]$
\State $\widetilde{Z}_T \leftarrow {\epsilon}$
\State $z \leftarrow f_{\theta}(\widetilde{Z}_T, T | u)$
\State Initialize \( \text{max\_score} \leftarrow -\infty \)
\State Initialize \( Z_{\text{best}} \leftarrow \text{null} \)
\For{$n = 1$ to $N - 1$}
    \State Sample ${\epsilon}_x \sim \mathcal{N}(0, I)$ and ${\epsilon}_h \sim \mathcal{N}(0, I)$
    \State Subtract center of gravity from ${\epsilon}_x$
    \State ${\epsilon} \leftarrow [\epsilon_x, \epsilon_h]$
    \State $\widetilde{Z}_{\tau_n} \leftarrow z + \sqrt{\tau_n - \tau_{n+1}} \cdot \epsilon$
    \State $z \leftarrow f_{\theta}(\widetilde{Z}_{\tau_n}, \tau_n | u)$
    \If{$n \geq m$}
        \State $ \text{score} \leftarrow \text{CustomScore}(z) $
        \If{$ \text{score} > \text{max\_score} $}
            \State $ \text{max\_score} \leftarrow \text{score} $
            \State $ Z_{\text{best}} \leftarrow z $
        \EndIf
    \EndIf
\EndFor
\State \textbf{output:} $ Z_{\text{best}} $
\end{algorithmic}
\end{algorithm}

\begin{algorithm}[ht]
\caption{Policy Gradient Version of RLCM}\label{alg:pg-rlcm}
\begin{algorithmic}[1]
\State \textbf{Input:} Consistency model policy $\pi_\theta = f_\theta(\cdot,\cdot) + \eps$, finetune horizon $H$, context set set $\Ccal$, batch size $b$, inference pipeline $P$
\For{$i = 1$ \textbf{to} $M$}
\State Sample $b$ contexts from $\Ccal$, $\bu \sim \Ccal$.
\State $\bm{Z} \gets P(f_\theta, H, \bu)$ \Comment{where $\bm{Z}$ is the batch of molecules}
\State Normalize rewards $r(\cdot, \cdot)$ per context
\State Split $\bm{Z}$ into $k$ minibatches.
\For{each minibatch}
\For{$t=0$ to $H$}
\State Accumulate gradients of $\theta$ using rule: 
\[
    \nabla_\theta \left[ \min\left\{ r(\bx_0,\bc) \cdot \frac{\pi_{\theta_{i+1}} (a_t| s_t)} {\pi_{\theta_{i}} (a_t | s_t)} , r(\bx_0,\bc) \cdot \texttt{clip} \left( \frac{\pi_{\theta_{i+1}} (a_t | s_t)} {\pi_{\theta_{i}} (a_t | s_t)}, 1-\eps, 1+\eps \right)   \right\} \right]
\]
\EndFor
\State Update parameters based on accumulated gradients.
\EndFor
\EndFor
\State Output trained consistency model $f_\theta(\cdot, \cdot)$
\end{algorithmic}
\end{algorithm}

\begin{algorithm}[ht]
\caption{TargetDiff/DecompDiff Inpainting Procedure}\label{alg:inpainting}
\begin{algorithmic}[4]
\State Set number of atoms to generate  based on reference scaffold. Get scaffold mask $m$.
\State Move center of mass of protein atoms to zero.
\State Sample initial molecular atom coordinates $x_T \sim \mathcal{N}(0, I)$
\State Sample initial bond types/atom types by $v_T = \text{one\_hot}(\arg\max_i g_i)$, where $g \sim \text{Gumbel}(0,1)$
\For{$t = T, \ldots, 1$}
    \For{$u = 1, \ldots, U$}
        \State 1) Get known 
        \State $x_{t-1}^{\text{known}} = \sqrt{\alpha_t} x_0 + (1 - \alpha_t) \epsilon$, where $\epsilon \sim \mathcal{N}(0, I)$
        \State $\log c = \log(\alpha_t v_0 + (1 - \alpha_t)/K)$
        \State $v_{t-1}^{\text{known}} = \text{one\_hot}(\arg\max_i [g_i + \log c_i])$, where $g \sim \text{Gumbel}(0,1)$
        \State 2) Get unknown 
        \State $x_{t-1}^{\text{unknown}} = \frac{1}{\sqrt{\alpha_t}} \left(x_t - \frac{\beta_t}{\sqrt{1-\alpha_t}} e_\theta(x_t, t)\right) + \sigma_t z$, $z \sim \mathcal{N}(0, I)$ 
        \State $v_{t-1}^{\text{unknown}}$ = Compute the posterior atom types $c(v_t, e_\theta(v_t, t))$
        \State 3) Merge according to mask $m$
        \State $x_{t-1} = m \odot x_{t-1}^{\text{known}} + (1 - m) \odot x_{t-1}^{\text{unknown}}$
        \State $v_{t-1} = m \odot v_{t-1}^{\text{known}} + (1 - m) \odot v_{t-1}^{\text{unknown}}$
        \If{$u < U$ and $t > 1$}
            \State Sample $x_{t}$ from the posterior $q_\theta(x_{t-1} \mid x_t)$
            \State Sample $v_{t}$ from the posterior $q_\theta(v_{t-1} \mid v_t)$
        \EndIf
    \EndFor
\EndFor
\State \Return $M$ with coordinates $x_0$ and atom types/ bond types $v_0$
\end{algorithmic}
\end{algorithm}

\clearpage 

\section{Information on Hyperparameters and Experiment Details}

\begin{table*}[ht]\centering
\raggedright{\textbf{Parameter setting for TurboHopp}}
\resizebox{\linewidth}{!}{
\begin{tabular}{p{0.5\linewidth}p{0.5\linewidth}}
\midrule[0.3ex]
\textbf{Setting} &
\textbf{Parameters} \\
\midrule[0.15ex]
\textbf{TurboHopp} \newline
Denoiser: GVP \newline
    \null\quad layers: 6 \newline
    \null\quad hidden features: 256\newline
    \null\quad GNN layers: 7 \newline
    \null\quad Attention: True \newline
    \null\quad Embedding size: 64 \newline
Optimizers: Adam \newline
    \null\quad $\gamma$: 1e-3 \newline
    \null\quad $\beta$: (0.9, 0.995)\newline
Dataset: PDBBind filtered \newline
Device: 4x NVIDIA A100 GPUs
& 
timesteps: 150, 100, 50, 25 \newline
batch size: 256 \newline
lr: 1e-4 \newline
schedule: ReduceLROnPlateau (min: 1e-6, factor: 0.9) \newline
num epochs: 5500 \newline
$\sigma_{\text{min}}$: 0.002 \newline
$\sigma_{\text{max}}$: 80.0 \newline
$\sigma_{\text{data}}$: 0.5 \newline
$\rho$: 7.0 \newline

\end{tabular}}
\end{table*}

\begin{table*}[ht!]
\raggedright{\textbf{Parameter setting for RLCM}}
\begin{tabular}{p{0.5\linewidth}p{0.5\linewidth}}
\midrule[0.3ex]
\textbf{Setting} &
\textbf{Parameters} \\
\midrule[0.15ex]
\textbf{Parameters for Docking Objective and Steric Clashes}\newline
Dataset: PDBBind filtered test set \newline
Device: 8x NVIDIA A100 GPUs \newline
& 
gradient accumulation steps: 1\newline
batch size: 215\newline
num epochs: 200\newline
sample iters: 1\newline
buffer size: 32\newline
min count: 16\newline
train batch size per gpu: 215\newline
num inner epochs: 1\newline
lr: 1e-5\newline
clip range: 1e-4\newline
max grad norm: 10
\end{tabular}

\end{table*}

\clearpage 
\section{RLCM and Consistency Model Training Curves}

For reproducibility, we present the curves from our training runs for both the consistency model and the use of RLCM.

\begin{figure}[ht!]
    \centering
    \includegraphics[width=\textwidth]{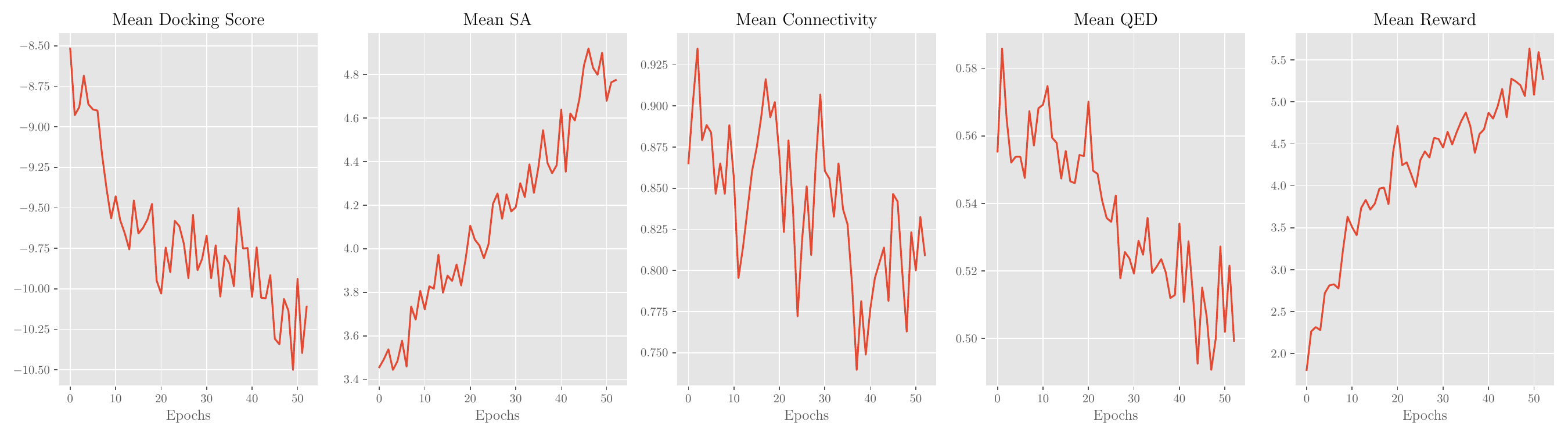}
    \caption{Training curves for the metrics which compose of the loss function. Notice that all either increase or maintain approximately the same value. Connectivity and QED score slightly decrease because we start from a previously RL finetuned checkpoint which optimizes only for connectivity, SA, and QED score.}
    \label{fig:rl}
\end{figure}

\begin{figure}[ht!]
    \centering
    \includegraphics[width=0.8\textwidth]{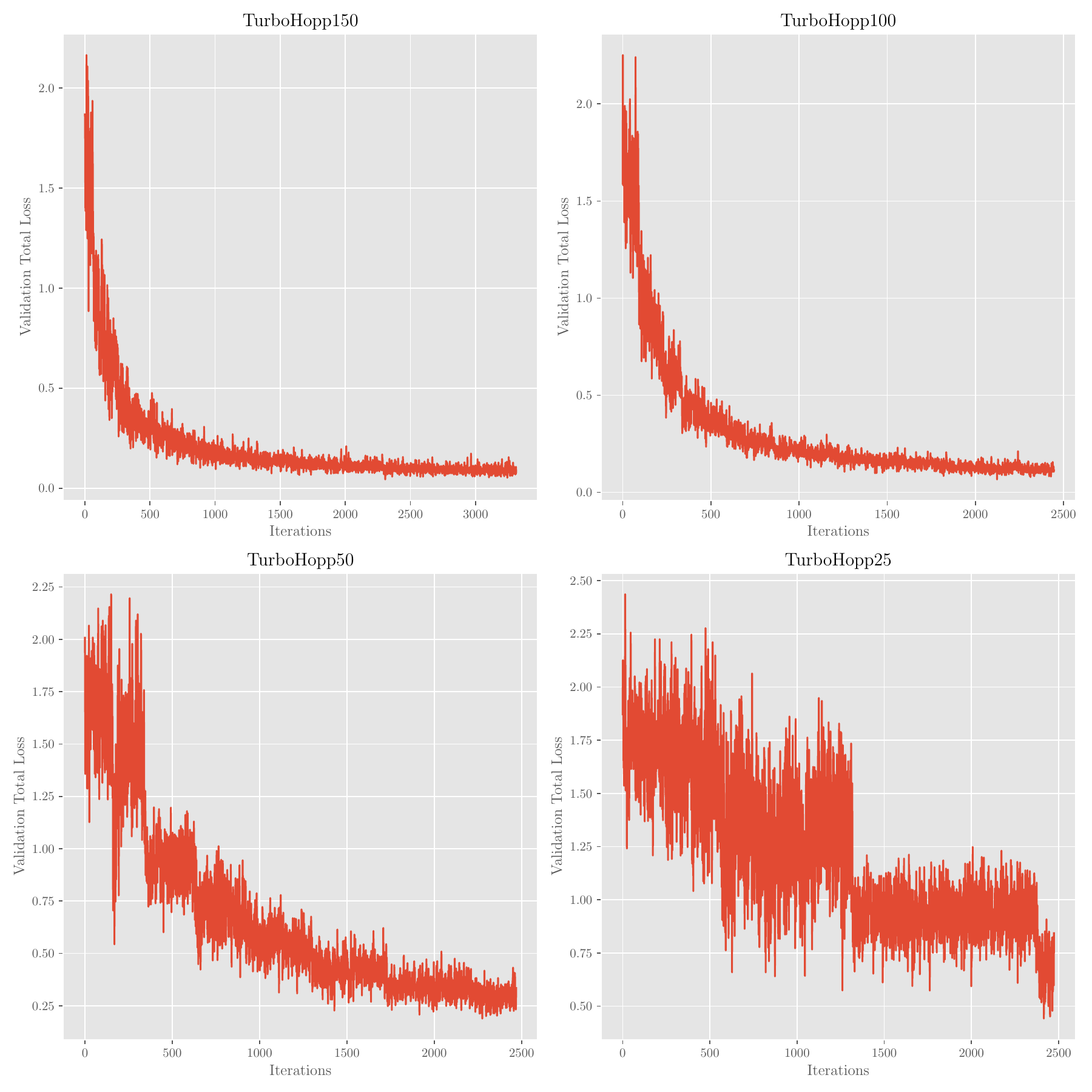}
    \caption{Training curves of total validation loss for different step size variants. We train a number of consistency models to empirically determine the optimal tradeoff between step size fidelity and speed. Turbohopp25 had low validity and proved to be too unstable for consistent generations.}
    \label{fig:cm_training}
\end{figure}

\begin{figure}[ht!]
    \centering
    \includegraphics[width=\textwidth]{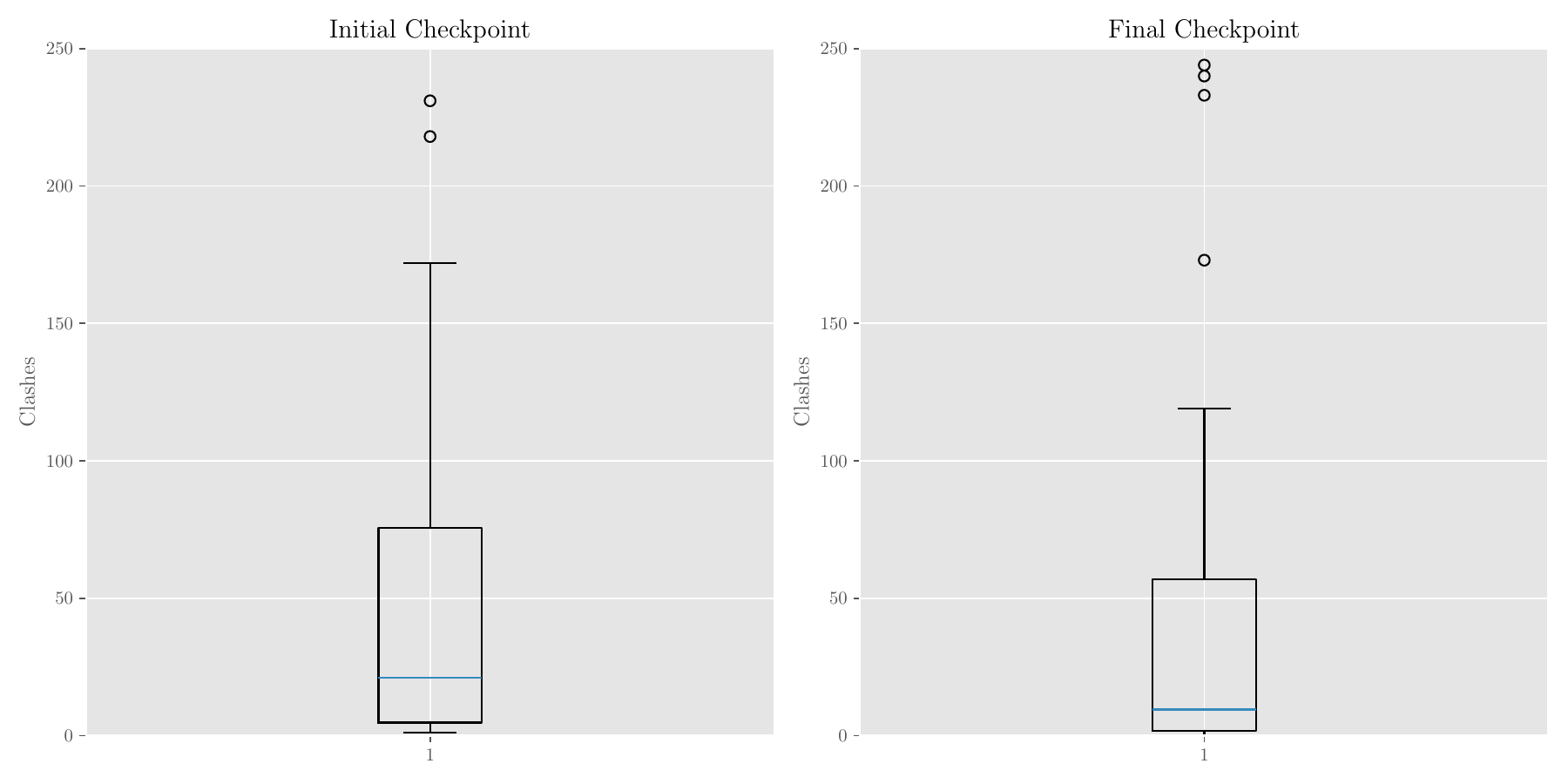}
    \caption{Plot of clashes before and after finetuning with a reward function mentioned in the main text. The initial Turbohopp-100 model was a RLCM finetuned model for connectivity, synthesizability, and QED score ($r_{\text{steric clash}}$. Notice the shift toward smaller number of clashes during training. However, we believe that further iteration of the reward function will lead to more effective finetuning but we leave this to future work.}
    \label{fig:turbohopp-clashes}
\end{figure}

\clearpage 
\section{Sample Molecules}

\begin{figure}[ht!]
\centering
    \includegraphics[width=1.0\textwidth]{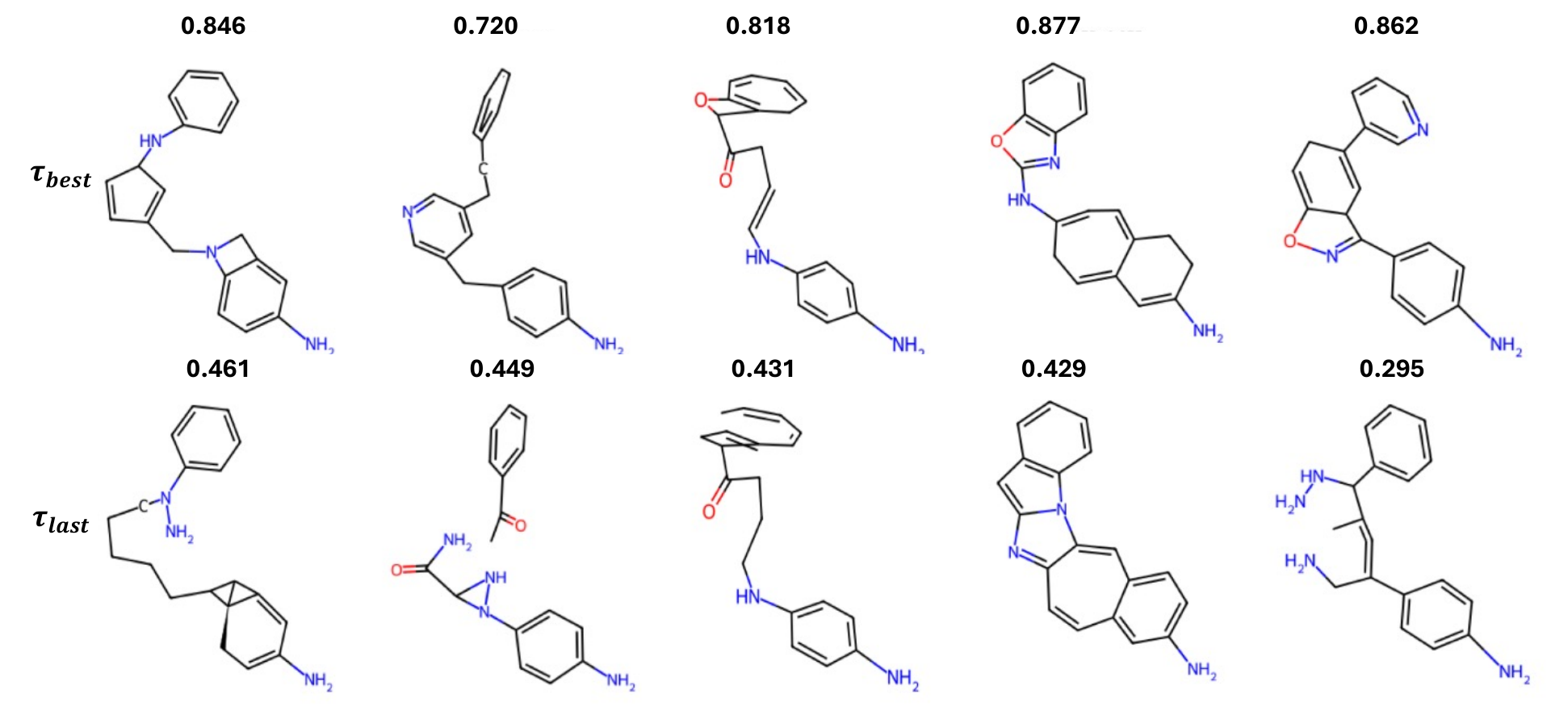}

    \caption{Examples of samples generated for PDB 6QQW with drug-likeliness. 1st row samples collected during multi-step phase with best scores and 2nd row indicates samples from the final step. Connectivity and overall metrics increased when we adopted custom score-based sampling.}
    \label{fig:turbohopp_mols}

\end{figure}

\begin{figure}[H]
  \centering
    \includegraphics[width=1.0\textwidth]{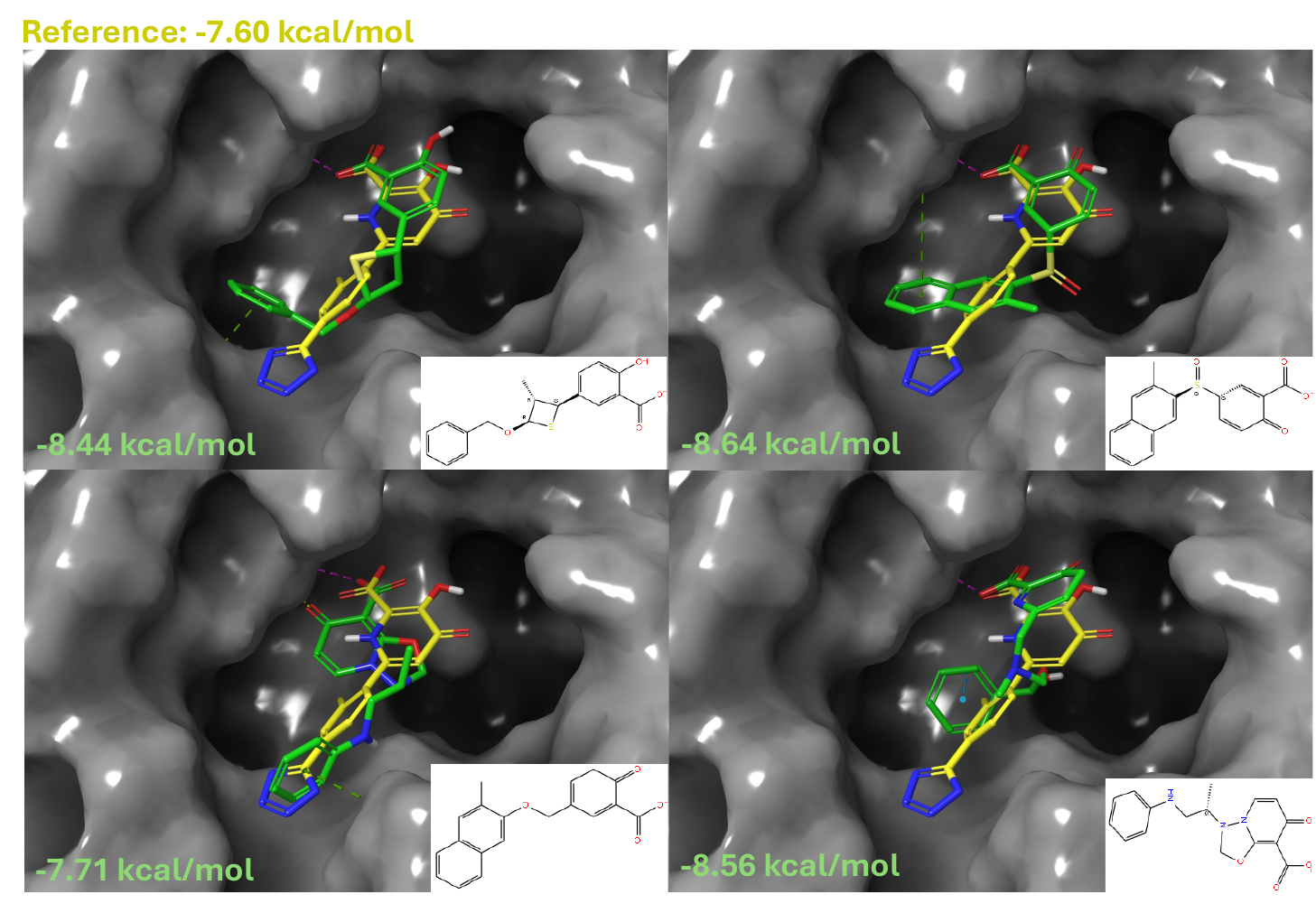}
  
    \caption{Scaffolds generated for PDB ID 6E6W by TurboHopp-100. Reference molecule is in yellow, while generated molecules are in green. Functional groups are the carboxylic acid and hydroxyl groups in the upper corner. Dotted lines in refer to ligand-receptor intermolecular interactions: green, blue, yellow, purple  being pi-cation, pi-pi stacking, hydrogen bonds, halogen bonds respectively. Compared to reference molecule, generated molecules had new interactions and higher binding affinity, while maintaining similar binding pose.}
\end{figure}

\begin{figure}[H]
    \centering
    \includegraphics[width=1.0\textwidth]{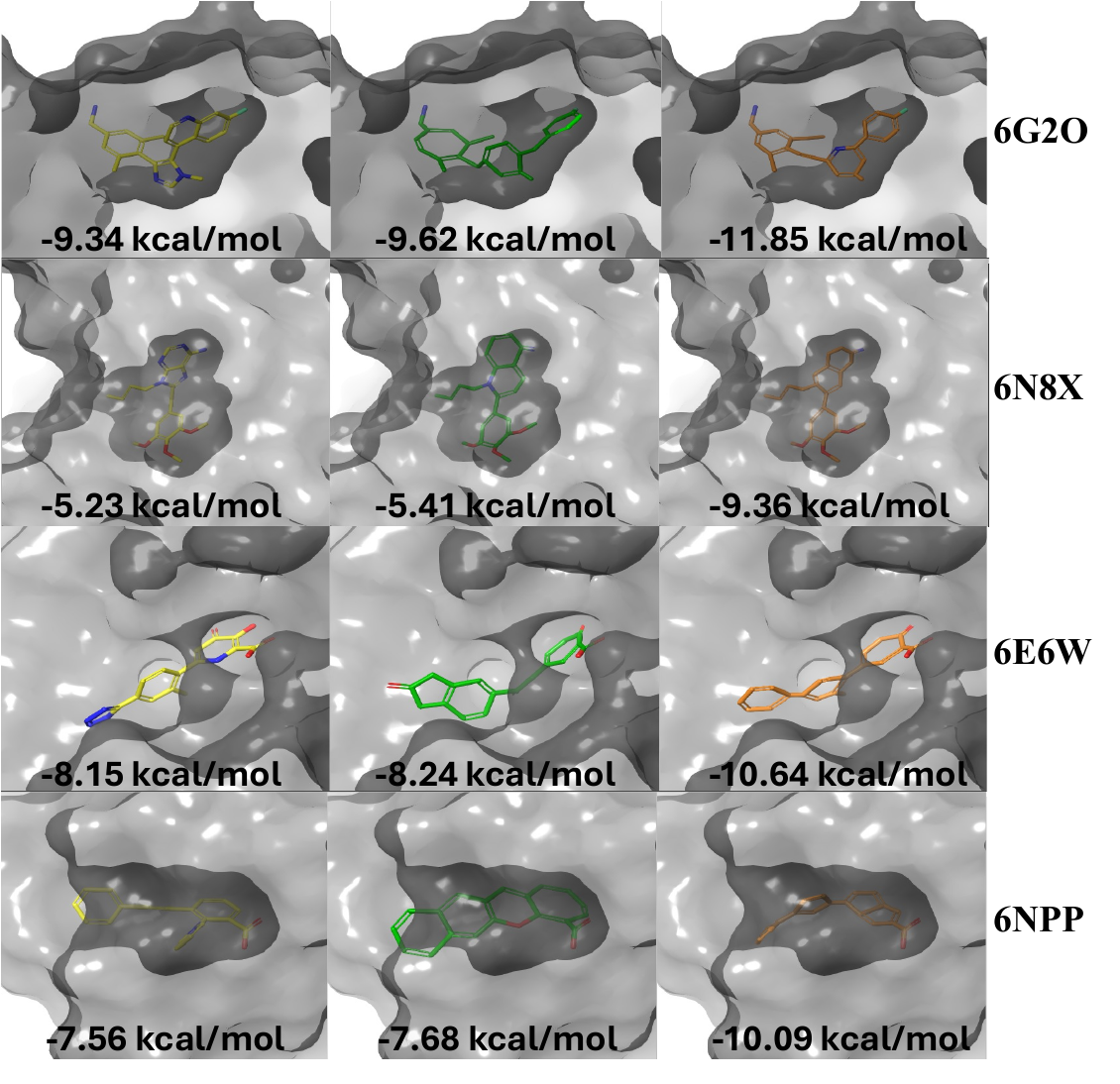}
    \caption{Comparison of Reference molecule(Yellow), and molecules generated by TurboHopp(Green) and TurboHopp-RL(Orange). Notice that Turbohopp and TurboHopp-RL generate molecules that have higher binding affinity with the protein.}
    \label{fig:docking-examples}
\end{figure}

\clearpage 
\begin{figure}[H]
  \centering
    \includegraphics[width=1.0\textwidth]{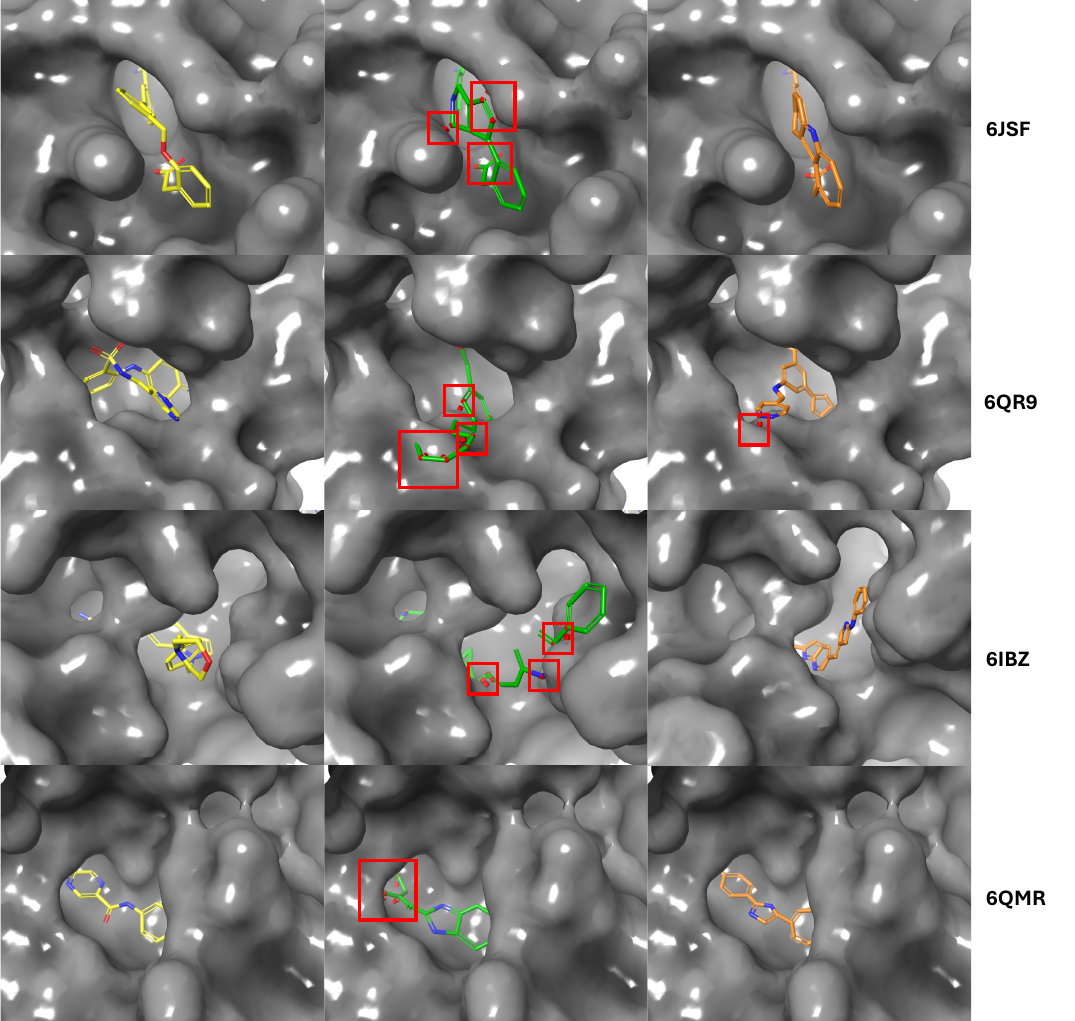}
  
    \caption{Comparison of Reference molecule(Yellow), and molecules generated by TurboHopp(Green) and TurboHopp-RL(Orange). Red box indicates collision points with protein atoms. TurboHopp-RL generates molecules that has less clashes with the protein. }
    \label{fig:turbohopp-clashes_example}
\end{figure}

\clearpage 
\section{Geometric Properties Comparison between TurboHopp and 3D-SBDD inpainting models}
\label{alg:geometric_props}
\begin{table}[ht]

\centering %
\caption{Jensen-Shannon divergence between bond distance distributions of generated VS reference(CrossDocked)}
 \vspace{10pt} 

\label{tab:comparison}
\begin{tabular}{@{}lcccc@{}} %
\toprule %
\textbf{Bond} & \textbf{TargetDiff\textsubscript{inpainting}} & \textbf{DecompDiff\textsubscript{inpainting}} & \textbf{TurboHopp-100} \\ %
\midrule %
C--C & 0.4984 & \textbf{0.2516} & 0.4376  \\
C=C & 0.6303 & \textbf{0.2962} & 0.4743  \\
C--N & 0.4833 & \textbf{0.2188} & 0.4178  \\
C=N & 0.6300 & \textbf{0.3913} & 0.5339  \\
C--O & 0.3882 & \textbf{0.2841} & 0.3809 \\
C=O & 0.3953 & \textbf{0.2395} & 0.3552  \\
C:C & 0.6829 & \textbf{0.2058} & 0.5562  \\
C:N & 0.6705 & \textbf{0.2734} & 0.5743  \\
\bottomrule %
\end{tabular}
\end{table}

\begin{table}[ht]

\centering %
\caption{Jensen-Shannon divergence between atom-atom distance distributions of generated VS reference(CrossDocked)}
 \vspace{10pt} 

\label{tab:comparison}
\begin{tabular}{@{}lcccc@{}} %
\toprule %
\textbf{Bond} & \textbf{TargetDiff\textsubscript{inpainting}} & \textbf{DecompDiff\textsubscript{inpainting}} & \textbf{TurboHopp-100} \\ %
\midrule %
C-C & 0.55 & \textbf{0.28} & 0.45  \\
all & 0.17 & \textbf{0.08} & 0.13  \\

\bottomrule %
\end{tabular}
\end{table}

\begin{table}[ht]
\centering
\caption{Percentage of different ring sizes for reference(CrossDocked) and model-generated molecules.}
\vspace{10pt}
\label{tab:ringSizeDistribution}
\begin{tabular}{ccccc}
\toprule
\textbf{Ring Size} & \textbf{Ref.} & \textbf{TargetDiff\textsubscript{inpainting}} & \textbf{DecompDiff\textsubscript{inpainting}} & \textbf{TurboHopp-100} \\
\midrule
3 & 1.70& 4.77& 0.00 & 7.85 \\
4 & 0.00& 1.12& 4.00& 0.81\\
5 & 30.20& 30.08& 37.21& 21.16\\
6 & 67.40& 57.91& 41.70 & 62.51 \\
7 & 0.70& 5.28& 14.06 & 6.19 \\
8 & 0.00& 0.81& 2.54 & 1.12 \\
9 & 0.00& 0.05& 0.48 & 0.31 \\
\bottomrule
\end{tabular}
\end{table}

\begin{table}[H]
\centering
\caption{Jensen-Shannon Divergence of Top 3 Torsion Angle Distributions (CrossDocked)}
\label{tab:torsion_angle_jsd}
\begin{tabular}{lccc}
\hline
\textbf{Metric} & \textbf{DecompDiff\textsubscript{inpainting}} & \textbf{TargetDiff\textsubscript{inpainting}} & \textbf{TurboHopp} \\ \hline
CCNC       & 0.184              & 0.204               & 0.207              \\
CCCC       & 0.264              & 0.249               & 0.197              \\
C:C:C:C    & 0.178              & 0.258               & 0.383              \\ \hline
\end{tabular}
\end{table}

\begin{table}[H]
\centering
\caption{Jensen-Shannon Divergence of Top 5 Bond Angle Distributions (CrossDocked)}
\label{tab:bond_angle_jsd}
\begin{tabular}{lccc}
\hline
\textbf{Metric} & \textbf{DecompDiff\textsubscript{inpainting}} & \textbf{TargetDiff\textsubscript{inpainting}} & \textbf{TurboHopp} \\ \hline
CNC        & 0.208              & 0.274               & 0.321              \\
CCN        & 0.168              & 0.207               & 0.332              \\
C:C:C      & 0.096              & 0.120               & 0.324              \\
CCC        & 0.231              & 0.231               & 0.298              \\
CCO        & 0.229              & 0.263               & 0.319              \\ \hline
\end{tabular}
\end{table}

\vspace{1050px}

\end{document}